\documentclass[letterpaper, 10pt, conference]{ieeeconf}

\usepackage{amsmath}
\usepackage{bm}
\usepackage{amssymb}
\usepackage{graphicx}
\usepackage{subcaption}
\usepackage{color}
\usepackage[usenames,dvipsnames,svgnames,table]{xcolor}
\usepackage{tabularx}
\usepackage{multirow}
\usepackage{url}
\let\labelindent\relax
\usepackage{enumitem}   

\title{\LARGE \bf
LS-VO: Learning Dense Optical Subspace for Robust Visual Odometry Estimation
}

\author{Gabriele Costante$^{\dagger,1}$ and Thomas A. Ciarfuglia$^{\dagger,1}$
\thanks{*We gratefully acknowledge the support of NVIDIA Corporation with the donation of the \emph{Titan Xp} GPU used for this research.}%
\thanks{$\dagger$ The authors contributed equally to the work.}%
\thanks{$^{1}$Department of Engineering, University of Perugia, Italy \newline
        {\tt\small thomas.ciarfuglia@unipg.it, gabriele.costante@unipg.it}}%
}

\input{settings/def_macros.tex}

\begin{document}

\maketitle
\thispagestyle{empty}
\pagestyle{empty}


\begin{abstract}
	This work proposes a novel deep network architecture to solve the camera Ego-Motion estimation problem. A motion estimation network generally learns features similar to Optical Flow (OF) fields starting from sequences of images. This OF can be described by a lower dimensional latent space. Previous research has shown how to find linear approximations of this space. We propose to use an Auto-Encoder network to find a non-linear representation of the OF manifold. In addition, we propose to learn the latent space jointly with the estimation task, so that the learned OF features become a more robust description of the OF input. We call this novel architecture Latent Space Visual Odometry (LS-VO).
	The experiments show that LS-VO achieves a considerable increase in performances with respect to baselines, while the number of parameters of the estimation network only slightly increases. 
\end{abstract}

\section{Introduction} \label{sec:introduction}
	
Learning based Visual Odometry (L-VO) in the last few years has seen an increasing attention of the robotics community because of its desirable properties of robustness to image noise and camera calibration independence \cite{costante2016exploring}, mostly thanks to Convolutional Neural Networks (CNNs) representational power, which can complement current geometric solutions \cite{gomez2017learning}.
While current results are very promising, making these solutions easily applicable to different environments still presents challenges. One of them is that most of the approaches so far explored have not shown strong domain independence and suffer from high dataset bias, i.e. the performances considerably degrade when tested on sequences with motion dynamics and scene depth significantly different from the training data \cite{tommasi2015deeper}. In the context of L-VO this bias is expressed in different Optical Flow (OF) field distribution in training and test data, due to differences in scene depth and general motion of the camera sensor.


One possible explanation for the poor performances of learned methods on unseen contexts is that most current learning architectures try to extract both visual features and motion estimate as a single training problem, coupling the appearance and scene depth with the actual camera motion information contained in the OF input. 
Some works have addressed the problem with an unsupervised, or semi-supervised approach, trying to learn directly the motion representation and scene depth from some kind of frame-to-frame photometric error \cite{ummenhofer2017demon} \cite{zhou2017unsupervised} \cite{vijayanarasimhan2017sfm}. While very promising, these approaches are mainly devised for scene depth estimation and still fall short in terms of general performances on Ego-Motion estimation. 

At the same time, previous research has shown how OF fields have a bilinear dependence on motion and inverse scene depth \cite{roberts2014optical}. We suggest that this is the main reason for the low generalization properties shown by learned algorithms so far.  
Past research has shown that the high dimensional OF field, when scene depth can be considered locally constant, can be projected on a much lower dimensional linear space \cite{roberts2008memory} \cite{wulff2015efficient}. However, when these conditions do not hold, the OF field subspace exists but is highly non-linear. 

\begin{figure}[t!]
\centering
	\includegraphics[width=0.37\textwidth]{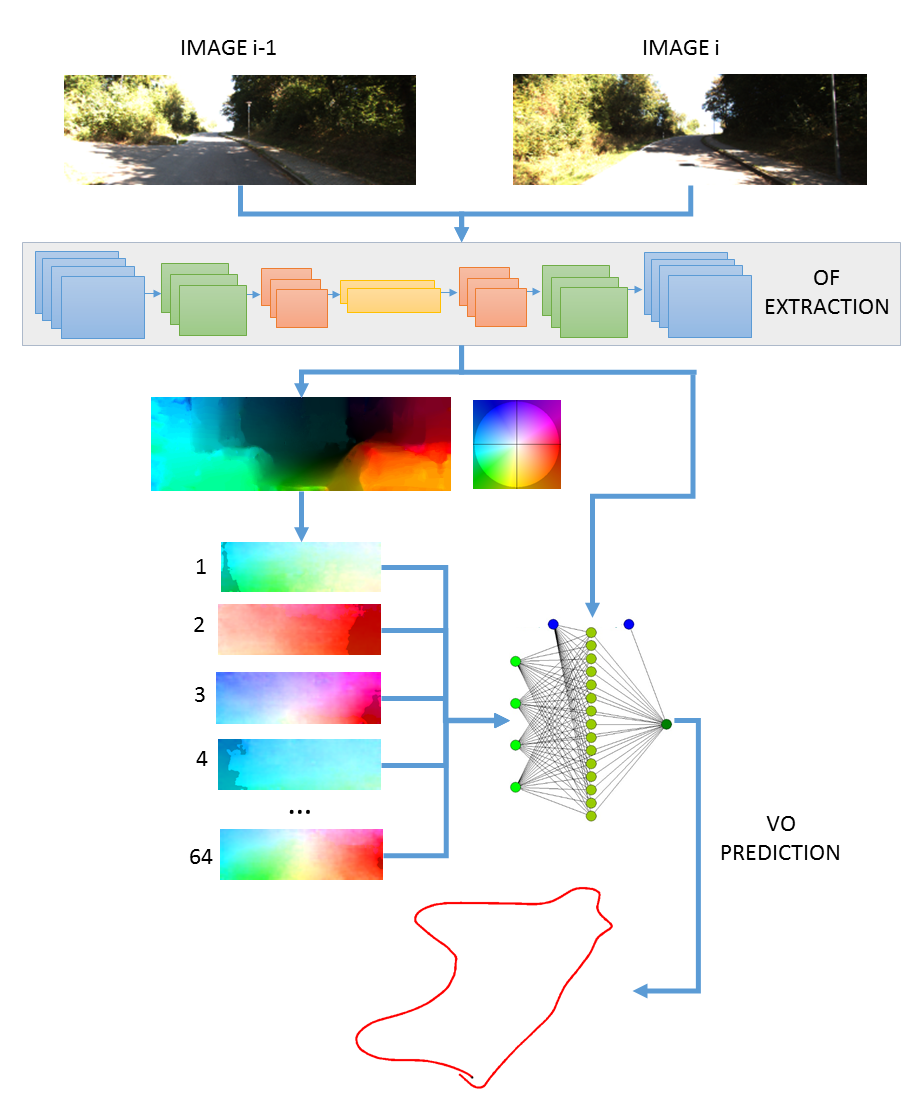} 
	\caption{\small Overview of the method: We propose a network architecture that jointly learn a latent space representation of the Optical Flow field and estimates motion. The joint learning makes the estimation more robust to input domain changes. The latent representation is an input to the estimation network together with the lower level features.\vspace{-1.4em}}
\label{fig:introduction}
\end{figure}

In this work we propose to exploit this knowledge, estimating the latent OF representation using an Auto-Encoder (AE) Neural Network architecture as a non-linear subspace approximator. AE networks are able to extract latent variable representation of high dimensional inputs. Since our aim is to make the Ego-Motion estimation more robust to OF fields that show high variability in their distribution, we do not simply use this subspace to directly produce motion prediction. Instead, we propose a novel architecture that jointly trains the subspace estimation and Ego-Motion estimation so that the two network tasks are mutually reinforcing and at the same time able to better generalize OF field representation. The conceptual architecture is shown in Figure \ref{fig:introduction}.
To demonstrate the increased performances and reduced dataset bias with respect to high dynamical variation of the OF field, we test the proposed approach on a challenging scenario. We sub-sample the datasets, producing sequences that simulate high speed variations, then we train and test on sequences that are both different in appearance and sub-sampling rate.

\section{Related Works} \label{sec:related}

\subsection{Ego-Motion estimation}

\subsubsection{Geometric Visual Odometry} \label{sec:related-gvo}
G-VO has a long history of solutions. While the first approaches were based on sparse feature tracking, mainly for computational reasons, nowadays direct or semi-direct approaches are preferred. These approaches use the photometric error as an optimization objective. Research on this topic is very active. Engel at al.  developed one of the most succesful direct approaches, LSD SLAM, both for monocular and stereoscopic cameras \cite{engel2014lsd}, \cite{engel2015large}. Forster et al. developed the Semi-Direct VO (SVO) \cite{forster2014svo} and its more recent update \cite{forster2017svo}, which is a direct method but tracks only a subset of features on the image and runs at very high frame rate compared to full direct methods.  Even if direct methods have gained most of the attention in the last few years, the ORB-SLAM algorithm by Mur-Artal et al. \cite{mur2015orb} reverted to sparse feature tracking and reached impressive robustness and accuracy comparable with direct approaches.

\subsubsection{Learned Visual Odometry} \label{sec:related-lvo}
Learned approaches go back to the early explorations by Roberts et al. \cite{roberts2008memory,roberts2009learning}, Guizilini et al. \cite{guizilini2011visual,guizilini2012semiparametric}, and Ciarfuglia et al. \cite{ciarfuglia2014evaluation}. As for the geometric case, the initial proposal focused on sparse OF features that, faithful to the \textit{there's no free lunch theorem}, explored the performances of different learning algorithms such as SVMs, Gaussian Processes and others. While these early approaches already showed some of the strengths of L-VO, it was only more recently, when Costante et al. \cite{costante2016exploring} introduced the use of CNNs for feature extraction from dense optical flow, that the learned methods started to attract more interest. Since then a couple of methods have been proposed. Muller and Savakis \cite{muller2017flowdometry} added the FlowNet architecture to the estimation network, producing one of the first end-to-end approaches. 
Clark et al. \cite{clark2017vinet} proposed an end-to-end approach that merged camera inputs with IMU readings using an LSTM network. Through this sensor fusion, the resulting algorithm is able to give good results but requires sensors other than a single monocular camera. 
The use of LSTM is further explored by Wang et al. in \cite{wang2017deepvo}, this time without any sensor fusion. The resulting architecture gives again good performances on KITTI sequences but does not show any experiments on environments with different appearance from the training sequences. 
On a different track is the work of Pillai et al. \cite{pillai2017towards}, that, like \cite{guizilini2012semiparametric}, looked at the problem as a generative probabilistic problem. Pillai proposes an architecture based on an MDN network and a Variational Auto-Encoder (VAE) to estimate the motion density given the OF inputs as a GMM. While Frame to Frame (F2F) performances are on a par with other approaches, they also introduce a loss term on the whole trajectory that mimics the bundle optimization that is often used in G-VO. The results of the complete system are thus very good. However, they use as input sparse KLT optical flow, since the joint density estimation for dense OF would become computationally intractable, meaning that they could be more prone to OF noise than dense methods.  

Most of the described approaches claim independence from camera parameters. While this is true, we note that this is more an intrinsic feature of the learning approach than the merit of a particular architecture. The learned model implicitly learns also the camera parameters, but then it fails on images collected with other camera optics. This parameter generalization issue remains an open problem for L-VO.

\subsection{Semi-supervised Approaches}\label{sec:related-semisupervised}
Since dataset bias and domain independence are critical challenges for L-VO, it is not surprising that a number of unsupervised and semi-supervised methods have been recently proposed. However, all the architectures have been proposed as a way of solving the more general problem of joint scene depth and motion estimation, and motion estimation is considered more as a way of improving depth estimation. Konda and Memisevich \cite{konda2015learning} used a stereo pair to learn VO but the architecture was conceived only for stereo cameras. Ummenhofer and Zhou \cite{ummenhofer2017demon} propose the DeMoN architecture, a solution for F2F Structure from Motion (SfM) that trains a network end-to-end on image pairs, levering motion parallax. 
Zhou et al. \cite{zhou2017unsupervised} proposed an end-to-end unsupervised system based on a loss that minimizes image warping error from one frame to the next. A similar approach is used by Vijayanarasimhan et al. \cite{vijayanarasimhan2017sfm} with their SfM-Net. 
All these approaches are devised mainly for depth estimation and the authors give little or no attention to the performances on VO tasks. Nonetheless, the semi-supervised approach is one of the more relevant future directions for achieving domain independence for L-VO, and we expect that this approach will be integrated in the current research on this topic.  
 
\subsection{Optical Flow Latent Space Estimation}\label{sec:related-subspaces}
The semi-supervised approaches described in Section \ref{sec:related-semisupervised} make evident an intrinsic aspect of monocular camera motion estimation, that is, even when the scene is static, the OF field depends both on camera motion and scene depth. This relationship between inverse depth and motion is bilinear and well known \cite{heeger1992subspace} and is at the root of scale ambiguity in monocular VO. However, locally and under certain hypothesis of depth regularity, it is possible to express the OF field in terms of a linear subspace of OF basis vectors. Roberts et al. \cite{roberts2009learning} used Probabilistic-PCA to learn a lower dimensional dense OF subspace without supervision, then used it to compute dense OF templates starting from sparse optical flow. They then used it to compute Ego-Motion. Herdtweck and Crist{\'o}bal extended the result and used Expert Systems to estimate motion  \cite{herdtweck2012experts}. More recently, a similar approach to OF field computation was proposed by Wulff and Black \cite{wulff2015efficient} that complemented the PCA with MRF, while Ochs et al.  \cite{ochs2017learning} did the same by including prior knowledge with an MAP approach. 
These methods suggest that OF field, which is an intrinsically high dimensional space generated from a non-linear process, lies on an ideal lower dimensional manifold that sometimes can be linearly locally approximated. However, modern deep networks are able to find latent representation of high dimensional image inputs, and in this work we use this intuition to explore this OF latent space estimation.

\section{Contribution} \label{sec:contribution}

Inspired by the early work of Roberts on OF subspaces \cite{roberts2014optical}, and by recent advances in deep latent space learning \cite{goodfellow2016deeplearningbook}, we propose a network architecture that jointly estimates a low dimensional representation of dense OF field using an Auto-Encoder (AE) and at the same time computes the camera Ego-Motion estimate with a standard Convolutional network, as in \cite{costante2016exploring}. The two networks share the feature representation in the decoder part of the AE, and this constrains the training process to learn features that are compatible with a general latent subspace. We show through experiments that this joint training increases the Ego-Motion estimation performances and generalization properties. 
In particular, we show that learning the latent space and concatenating it to the feature vector makes the resulting estimation considerably more robust to domain change, both in appearance and in OF field dynamical range and distribution. 

We train our network both in an end-to-end version, using deep OF estimation, and with standard OF field input, in order to explore the relative advantages and weaknesses. We show that while the end-to-end approach is more general, precomputed OF still has some performance advantages. 

In summary our contributions are:
\begin{itemize}
\item A novel end-to-end architecture to jointly learn the OF latent space and camera Ego-Motion estimation is proposed. We call this architecture Latent Space-VO (LS-VO).
\item The strength of the proposed architecture is demonstrated experimentally, both for appearance changes, blur, and large camera speed changes. 
\item Effects of geometrically computed OF fields are compared to end-to-end architectures in all cases. 
\item The adaptability of the proposed approach to other end-to-end architectures is demonstrated, without increasing the chances of overfitting them, due to parameters increase.

\end{itemize}

\begin{figure*}[ht!]
\centering
	\includegraphics[scale=0.2]{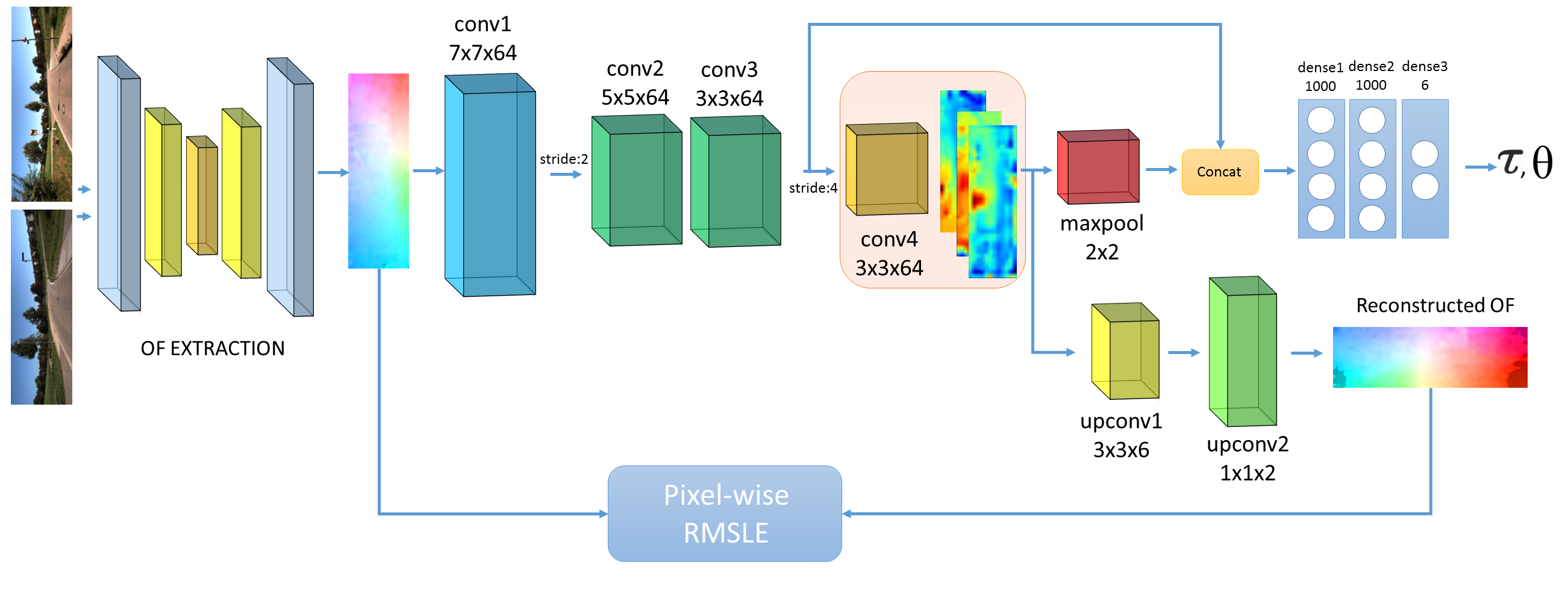} 
	\vspace{-1.4em}
	\caption{\small  LS-VO network architecture. The shared part is composed of Flownet OF extraction, then three convolutional layers that start the feature extraction. The last layer of the Encoder, conv4, is not shared with the Estimator network. From conv4 the latent variables are produced. The Decoder network takes these variables and reconstructs the input, while the Estimator concatenates them to conv3 output. Then three fully connected layers produce the motion estimates.\vspace{-1.4em}} 
	\label{fig:network-architecture}
\end{figure*}

\section{Learning Optical Flow Subspaces} \label{sec:approach}

Given an optical flow vector $\mbf{u} = \T{(\T{\ub}_x, \T{\ub}_y)}$ from a given OF field $\xb$, \cite{roberts2014optical} \cite{wulff2015efficient} approximate it with a linear relationship: 
\begin{equation}
 \ub \approx \mbf{W} \zb = \sum_{i=1}^{l} z_i \mbf{w}_i 
\end{equation}
where the columns of $\mbf{W}$ are the basis vectors that form the OF linear subspace and $\zb$ is a vector of latent variables. This approximation is valid only if there are some regularities of scene depth and is applicable only to local patches in the image. The real subspace is non-linear in nature and, in this work, we express it as a generic function $\ub = \mathcal{D}(\zb )$ that we learn from data by using the architecture described in the following.

\subsection{Latent Space Estimation with Auto-Encoder Networks} \label{sec:approach-overview}
Let $\yb \in \Rbb^6$ be the camera motion vector and $\xb \in \Rbb^{2 \times w \times h}$ the input OF field, computed with some dense method, where $\mathit{\xb}_{(i,j)} = \ub_{(i,j)}$ is a 2-dimensional vector of the field at image coordinates $(i,j)$. Both can be viewed as random variables with their own distributions. In particular, we make the hypothesis that the input images lie on a lower dimensional manifold, as in \cite{zhu2016generative}, and thus also the OF field lies on a lower dimensional space $\Obb \subset \Rbb^{2 \times w \times h}$ with a distance function $S(\xb^{(a)}, \xb^{(b)})$, where $\xb^{(a)}, \xb^{(b)} \in \mathbb{O}$. The true manifold is very difficult to compute, so we look for an estimate $\hat{\mathbb{O}} \approx \mathbb{O}$ using the model extracted by an encoding neural network. 

Let $\zb \in \Ybb \subset \Rbb^l, l \ll w \times h$ be a vector of latent random variables that encodes the variabilities of OF field that lies on this approximate space. 
The decoder part of the AE can be seen as a function 
\begin{equation}
	\mathcal{D}(\zb; \theta_d) = D(\zb; \{\mbf{W}_k, \mbf{b}_k\}, k=1 \cdots K) 
\end{equation}
where $\theta_d=(\{\mbf{W}_k, \mbf{b}_k\}, k=1 \cdots K)$ is the set of learnable parameters of the network (with $K$ upconv layers), that is able to generate a dense optical flow from a vector of latent variables $\zb$. 
Note that
the AE works similarly to a non-linear version of PCA \cite{goodfellow2016deeplearningbook}. We define the set $\hat{\Obb} = \left\lbrace D(\zb; \theta_d) \; | \; \zb \in \Ybb \right\rbrace$ as our approximation of the OF field manifold and use the logaritmic Euclidean distance (as described in Section \ref{sec:approach-architecture} as a loss function) as an approximation of $S(D(\zb^{(a)}),D(\zb^{(b)}))$. Using this framework the problem of estimating the latent space is carried out by the AE network, where the Encoder part can be defined as the function $\zb = E(\xb; \theta_e)$. 

While in \cite{pillai2017towards} the AE is used to estimate motion, and $\zb$ are the camera translation and rotations, here we follow a different strategy. 
We compute the latent space for a two-fold purpose: we use the latent variables as an input feature to the motion estimation network and we learn this latent space together with the estimator, thus forcing the estimator to learn features compatible with the encoder representation. Together these two aspects make the representation more robust to domain changes.

\subsection{Network Architecture} \label{sec:approach-architecture}
The LS-VO network architecture in its end-to-end form is shown in Figure \ref{fig:network-architecture}. It is composed of two main branches, one is the AE network and the other is the convolutional network that computes the regression of motion vector $\yb$.  The OF extraction section is Flownet \cite{fischer2015flownet}, for which we use the pre-trained weights. We run tests fine-tuning this part of the network on KITTI \cite{geiger2013vision} and Malaga \cite{blanco2013mlgdataset} datasets, but the result was a degraded performance due to overfitting. 

The next layers are convolutions that extract features from the computed OF field. After the first convolutional layers (conv1, conv2 and conv3), the network splits into the AE network and the estimation network. The two branches share part of the feature extraction convolutions, so the entire network is constrained in learning a general representation that is good for estimation and latent variable extraction. The Encoder is completed by another convolutional layer, that brings the input $\xb$ to the desired representation $\zb$, and its output is fed both in the Decoder and concatenated to the feature extracted before. The resulting feature vector, composed of latent variables and convolutional features is fed into a fully connected network that performs motion estimation. The details are summarized in Table \ref{tab:network-param}. 

The AE is trained with a pixel-wise squared Root Mean Squared Log Error (RMSLE) loss:

\begin{equation}
\mathcal{L}_{AE} = \sum_i || \log(\hat{\ub}^{(i)} +\mbf{1}) - \log(\ub^{(i)} +\mbf{1}) ||_2^2
\end{equation}

where $\hat{\ub}^{(i)}$ is the predicted OF vector for the i-th pixel, and $\ub^{(i)}$ is the corresponding input to the network, and the logarithm is intended as an element-wise operation. This loss penalizes the ratio difference, and not the absolute value difference of the estimated OF compared to the real one, so that the flow vectors of distant points are taken into account and not smoothed off.

We use the loss introduced by Kendall et al. in \cite{Kendall_2015_ICCV}:
\begin{equation}
\mathcal{L}_{EM} = \sum_i || \hat{\boldsymbol{\tau}} - \boldsymbol{\tau} ||_2^2 + \beta || \hat{\boldsymbol{\theta}} - \boldsymbol{\theta} ||_2^2
\label{eq:kendall-loss}
\end{equation}
where the $\boldsymbol\tau$ is camera translation vector in meters, $\boldsymbol\theta$ is the rotation vector in Euler notation in radians, and $\beta$ is a scale factor that balances the angular and translational errors. $\beta$ has been cross-validated on the trajectory reconstruction error ($\beta=20$ for our experiments), so that the frame to frame error propagation to the whole trajectory is taken into account. The use of a Euclidean loss with Euler angle representation works well in the case of autonomous cars, since the yaw angle is the only one with significant changes. For more general cases, is better to use a quaternion distance metric \cite{Kuffner_2004_ICRA}.

\begin{table}[]
\caption{LS-VO and ST-VO network architectures }
\label{tab:network-param}

\centering


\begin{tabular}{|c||c|c|c|c|}

\cline{1-5}
{} & \textbf{Layer name} & \textbf{Kernel size} & \textbf{Stride} & \textbf{output size} \\
\hline
Input & - & - & - & $(94, 300, 2)$\\
\hline
\hline
{} & \multicolumn{4}{c|}{\textbf{LS-VO}}\\
\hline
\multirow{3}{1.2cm}{\centering Shared Features Layer} & conv1 & $7 \times 7 $ & $2 \times 2$ & $(47, 150, 64)$ \\
\cline{2-5}
{} & conv2 & $5 \times 5 $ & $1 \times 1$ & $(47, 150, 64)$ \\
\cline{2-5}
{} & conv3 $\star$ & $3 \times 3 $ & $4 \times 4$ & $(12, 38, 64)$ \\
\hline
\multirow{3}{1.2cm}{\centering Auto-Encoder} & conv4 & $3 \times 3 $ & $1 \times 1$ & $(12, 38, 64)$ \\
\cline{2-5}
{} & upconv1 & $3 \times 3 $ & $1 \times 1$ & $(48, 152, 6)$\\
\cline{2-5}
{} & crop & - & - & $(47, 150, 6)$ \\
\cline{2-5}
{} & upconv2 & $1 \times 1 $ & $1 \times 1$ & $(94, 300, 2)$ \\
\hline
\multirow{3}{1.2cm}{\centering Estimator} & maxpool $\dagger$ & $2 \times 2$ & $2 \times 2$ & $(6, 19, 64)$ \\
\cline{2-5}
{} & concat $\star$ and $\dagger$ & - & - & $(36480)$  \\
\cline{2-5}
{} & dense1 & - & - & $(1000)$ \\
\cline{2-5}
{} & dense2 & - & - & $(1000)$ \\
\cline{2-5}
{} & dense3 & - & - & $(6)$ \\

\hline
\hline

{} & \multicolumn{4}{c|}{\textbf{ST-VO}}\\
\hline
\multirow{4}{1.2cm}{\centering Feature Extraction} & st-conv1 & $3 \times 3 $ & $2 \times 2$ & $(46, 149, 64)$ \\
\cline{2-5}
{} & st-maxpool1 $\bullet$ & $4 \times 4 $ & $4 \times 4$ & $(11, 37, 64)$\\
\cline{2-5}
{} & st-conv2 & $4 \times 4 $ & $4 \times 4$ & $(9, 35, 20)$\\
\cline{2-5}
{} & st-maxpool2 $\diamond$ & $2 \times 2 $ & $2 \times 2$ & $(4, 17, 20)$ \\
\hline
\multirow{3}{1.2cm}{\centering Estimation} & concat $\bullet$ and $\diamond$ & - & - & $(27408)$ \\
\cline{2-5}
{} & st-dense1 & - & - & $(1000)$ \\
\cline{2-5}
{} & st-dense2 & - & - & $(6)$\\
\hline

\end{tabular}
\vspace{-1.4em}

\end{table}

In Section \ref{sec:results}, we compare this architecture both with SotA geometrical and learned methods. The baseline for the learned approaches is a Single Task (ST) network, similar to the 1b network presented in \cite{costante2016exploring}, and described in Table \ref{tab:network-param}.

\subsection{OF field distribution}\label{sec:approach-of-distribution}

\begin{figure}
\centering

	\begin{subfigure}[b]{0.5\textwidth}
	\centering
        \includegraphics[width=0.45\linewidth]{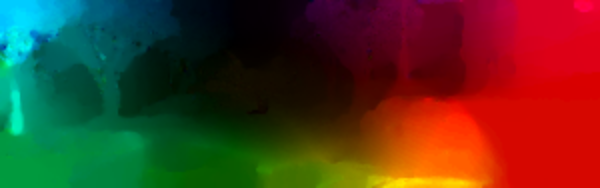}
        \includegraphics[width=0.45\linewidth]{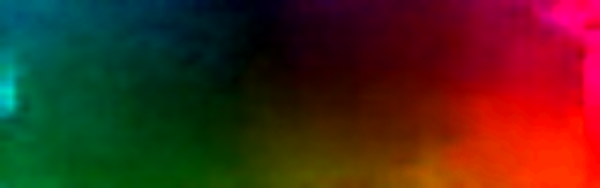}
        \caption{no sub-sampling - 10Hz}
        \vspace{1mm}
        \label{fig:of-dns1}
    \end{subfigure}
    
    \begin{subfigure}[b]{0.5\textwidth}
    \centering
        \includegraphics[width=0.45\linewidth]{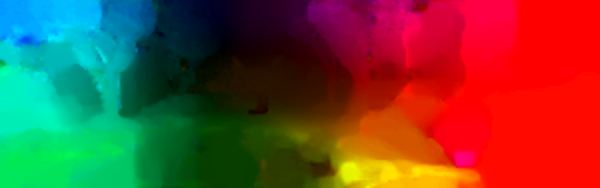}
        \includegraphics[width=0.45\linewidth]{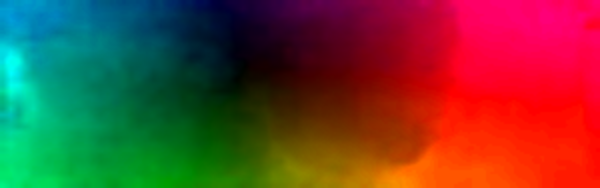}
        \caption{sub-sampling 2 - 5Hz}
        \vspace{1mm}
        \label{fig:of-dns2}
    \end{subfigure}
   
    \begin{subfigure}[b]{0.5\textwidth}
    \centering
        \includegraphics[width=0.45\linewidth]{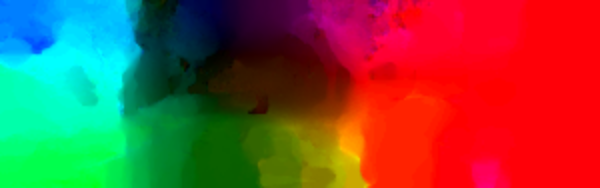}
        \includegraphics[width=0.45\linewidth]{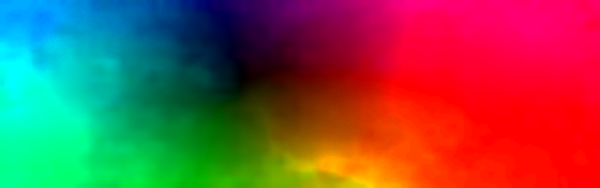}
        \caption{sub-sampling 3 - 3.33Hz}
        \vspace{1mm}
        \label{fig:of-dns3}
    \end{subfigure}
    
	\caption{\small Examples of the OF field intensity due to different sub-sampling rates of the original sequences. In the left are the OF field extracted with Brox algorithm (BF) \cite{brox2004high}, while on the right the ones extracted with Flownet \cite{fischer2015flownet}. While the BF fields look more crisp, they require parameter tuning, while the Flownet version is non-parametric at test time. \label{fig:of-comparison} }
\end{figure}

\begin{figure}
\centering

	\begin{subfigure}[b]{0.5\textwidth}
	\centering
        \includegraphics[width=0.45\linewidth]{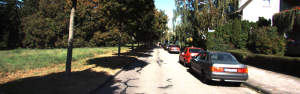}
        \includegraphics[width=0.45\linewidth]{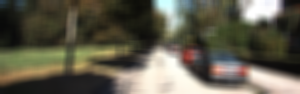}
        \caption{Standard and blurred image}
        \vspace{1mm}
        \label{fig:of-blur2}
    \end{subfigure}
    
    \begin{subfigure}[b]{0.5\textwidth}
    \centering
        \includegraphics[width=0.45\linewidth]{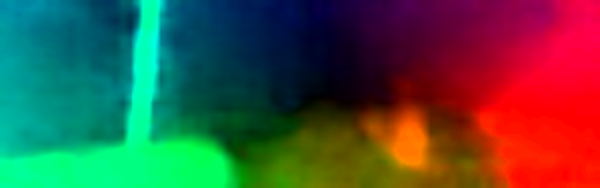}
        \includegraphics[width=0.45\linewidth]{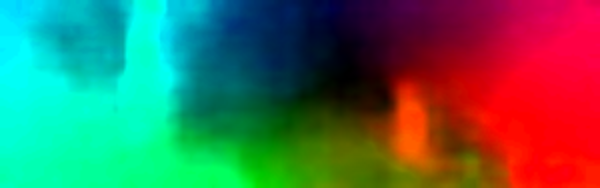}
        \caption{Standard and blurred OF field}
        \label{fig:of-blur2}
    \end{subfigure}
    
	\caption{\small Examples of OF fields obtained applying gaussian blur to image sequences. (a) The image and its blurred variant is shown, with blur radius 10. (b) The corresponding OF fields. Note the huge change in OF distribution. \label{fig:of-blur} \vspace{-1.7em}}
\end{figure}

As mentioned in Section \ref{sec:approach-overview}, the OF field has a probability distribution that lies on a manifold with lower dimensionality than the number of pixels of the image. We can argue that the actual density depends on the motion of the camera as much as the scene depth of the images collected. In this work, we test generalization properties of the network for both aspects: 
\begin{enumerate}[label=\roman*, ref=(\roman*)]
\item For the appearance we use the standard approach to test on completely different sequences than the ones used in training. \label{i-1}
\item For the motion dynamics, we sub-sample the sequences, thus multiplying the OF dynamics by the same factor. \label{i-2}
\item To further test OF distribution robustness, we also test the architecture on downsampled blurred images, as in \cite{costante2016exploring}. \label{i-3}
\end{enumerate}
Examples of the resulting OF field are shown in Figure \ref{fig:of-comparison}, while an example of a blurred OF field is shown in Figure \ref{fig:of-blur}. In both images there are evident differences both in hue and saturation, meaning that both modulus and phase of the OF vectors change. 


\section{Experimental Results}
\label{sec:results}

\subsection{Data and Experiments set-up}\label{sec:results-data}
We perform experiments on two different datasets, the KITTI Visual Odometry benchmark \cite{geiger2013vision} and the Malaga 2013 dataset \cite{blanco2013mlgdataset}. Both datasets are taken from cars that travel in city suburbs and countryside, however the illumination conditions and camera setups are different.
For the KITTI dataset we used the sequences $00$ to $07$ for training and the $08$, $09$ and $10$ for test, as is common practice. The images are all around $1240 \times 350$, and we resize them to $300 \times 94$. The frame rate is $10$Hz. 
For the Malaga dataset we use the sequences $02$, $03$ and $09$ as test set, and the $01$, $04$, $06$, $07$, $08$, $10$ and $11$ as training set. In this case the images are $1024 \times 768$ that we resize to $224 \times 170$. The frame rate is $20$Hz. For the Malaga dataset there is no high precision GPS ground truth, so we use the ORBSLAM2 stereo VO \cite{mur2015orb} as a Ground truth, since its performances, comprising bundle adjustment and loop closing, are much higher than any monocular method.

The networks are implemented in Keras/Tensorflow and trained using an Nvidia Titan Xp. Training of the ST-VO variant takes $6h$, while LS-VO $27h$.
The ST-VO memory occupancy is on average $460$MB, while LS-VO requires $600$MB. 
At test time, computing Flownet and BF features takes on average $12.5$ms and $1$ ms per sample, while the prediction requires, on average, $2-3$ms for both ST-VO and LS-VO. The total time, when considering Flownet features, amounts to $14.5$ms for ST-VO and $15.5$ms for LS-VO. Hence, we can observe that the increased complexity does not affect much computational performance at test time.

For all the experiments described in the following Section, we tested the LS-VO architecture and the ST-VO baseline. Furthermore, on all KITTI experiments we tested with both Flownet and BF features. While the contribution of this work relates mainly on showing the increased robustness of the proposed method with respect to learned architectures, we also sampled the performances of SotA geometrical methods, namely VISO2-M \cite{geiger2011stereoscan} and ORBSLAM2-M \cite{mur2015orb} in order to have a general baseline. 


\subsection{Experiments}\label{sec:results}
As mentioned in Section \ref{sec:approach-of-distribution}, on both datasets we perform three kinds of experiments, of increasing difficulty. We observe that the original sequences show some variability in speed, since the car travels in both datasets at speeds of up to $60$Km/h, but the distribution of OF field is still limited. This implies that the possible combinations of linear and rotational speeds are limited. We extend the variability of OF field distribution performing some data augmentation. Firstly, we sub-sample the sequences by 2 and 3 times, to generate virtual sequences that have OF vectors with very different intensity. In Figure \ref{fig:of-comparison}, an example of the different dynamics is shown. In both KITTI and Malaga datasets we indicate the standard sequences by the ${d1}$ subscript, and the sequences sub-sampled by $2$ and $3$ times by ${d2}$ and ${d3}$, respectively. In addition to this, we generate blurred versions of the ${d2}$ test sequences, with gaussian blur, as in \cite{costante2016exploring}. Then we perform three kinds of experiment and compare the results. The first is a standard training and test on $d1$ sequences. This kind of test explores the generalization properties on appearance changes alone. In the second kind of experiment we train all the networks on the sequences ${d1}$ and ${d3}$ and test on ${d2}$. This helps us to understand how the networks perform when both appearance and OF dynamics change. The third experiment is training on $d1$ and $d3$ sequences, and testing on the on the blurred versions of the $d2$ test set (Figure \ref{fig:of-blur}).
%

The proposed architecture is end-to-end, since it computes the OF field through a Flownet network. However, as a baseline, we decided to test the performances of all the architecture on a standard geometrical OF input, computed as in \cite{brox2004high}, and indicated as BF in the following. 

In addition, we train the BF version on the RGB representation of OF, since from our experiments performs slightly better than the floating point one.

\begin{figure}

	\centering
	\begin{subfigure}{0.45\linewidth}
		\includegraphics[width=\textwidth]{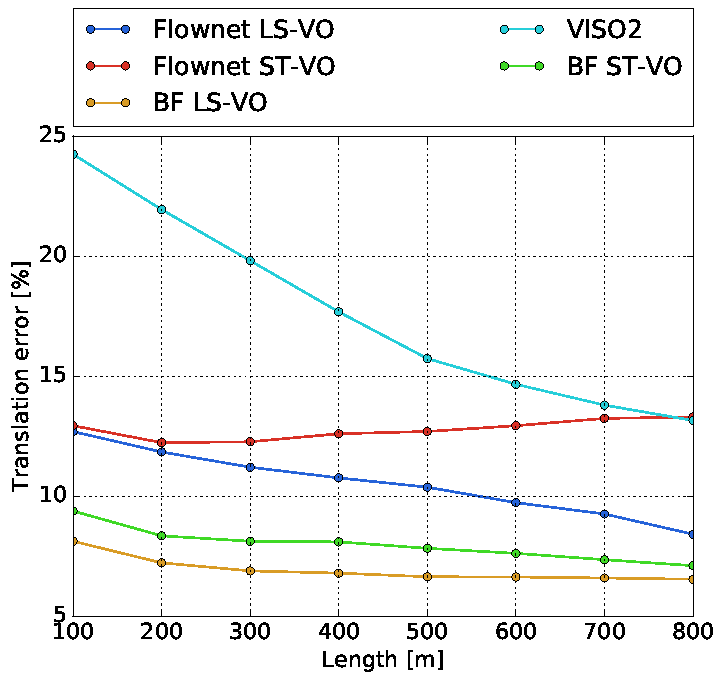}
		\caption{length vs tran. error}
		\label{fig:results-kitti-dns1-L-a}
	\end{subfigure}
	\begin{subfigure}{0.45\linewidth}
		\includegraphics[width=\textwidth]{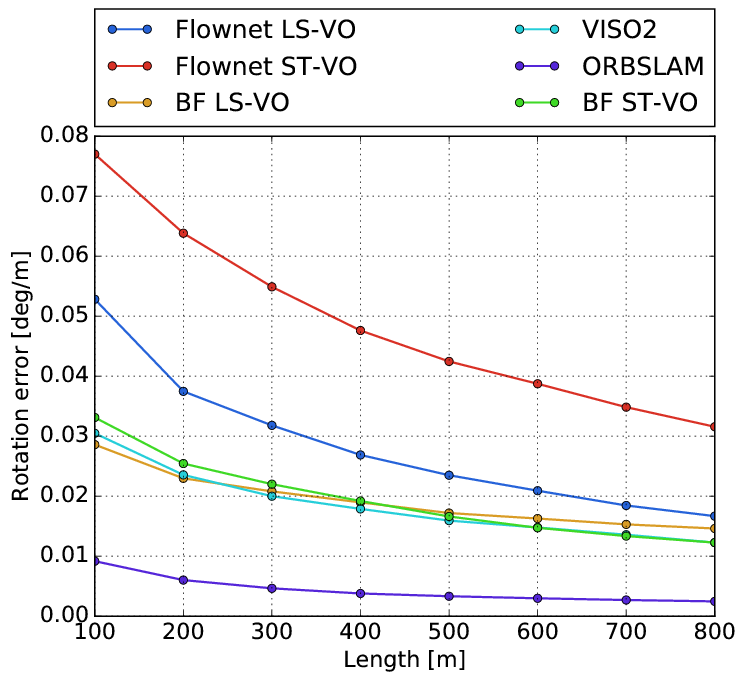}
		\caption{length vs rot. error}
		\label{fig:results-kitti-dns1-L-b}
	\end{subfigure}
	\begin{subfigure}{0.45\linewidth}
		\includegraphics[width=\textwidth]{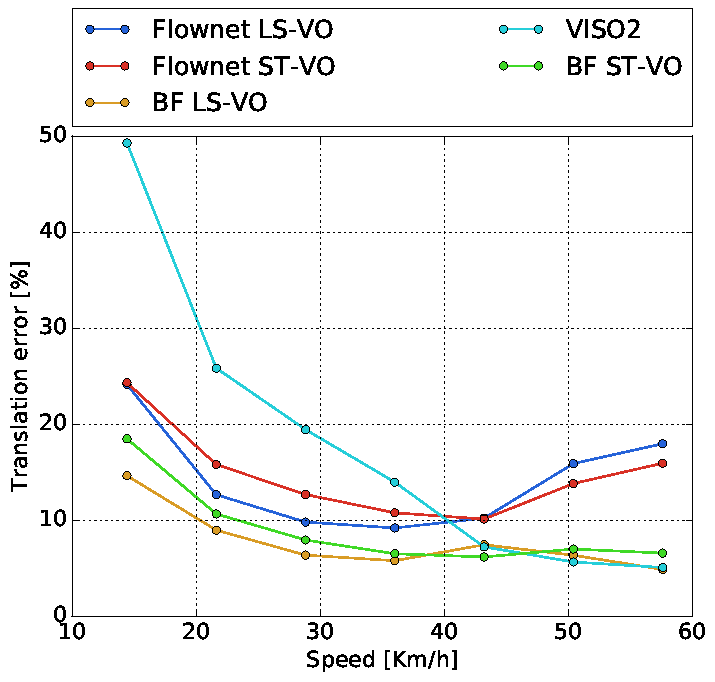}
		\caption{speed vs tran. error}
		\label{fig:results-kitti-dns1-L-c}
	\end{subfigure}
	\begin{subfigure}{0.45\linewidth}
		\includegraphics[width=\textwidth]{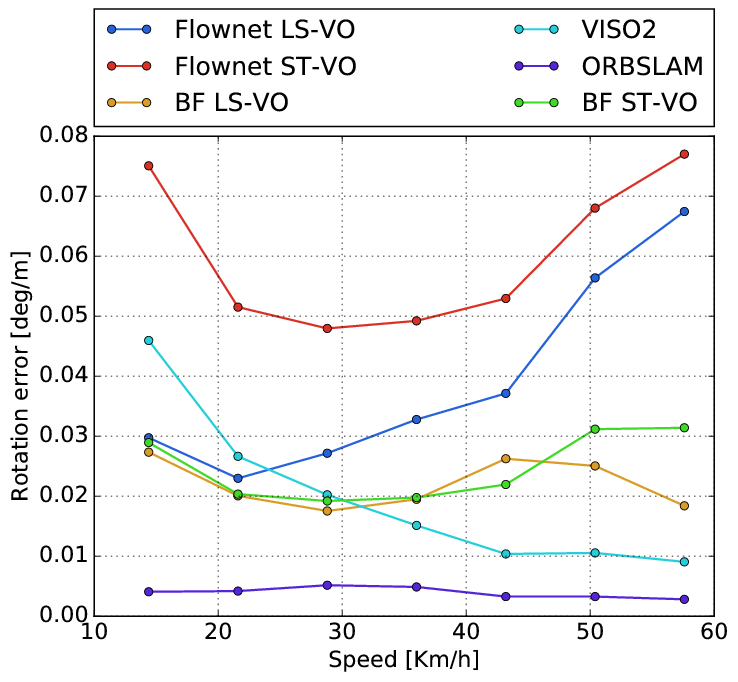}
		\caption{speed vs rot. error}
		\label{fig:results-kitti-dns1-L-d}
	\end{subfigure}
	\caption{\small Comparison between all methods on KITTI dataset, with no sequence sub-sampling. It is evident that the LS-VO network outperforms the ST equivalent, and in the case of the BF OF inputs it is almost always better by a large margin. Geometrical methods outperform learned ones on angular rate. ORBSLAM2-M is not shown in \ref{fig:results-kitti-dns1-L-a} and \ref{fig:results-kitti-dns1-L-b} for axis clarity, since the error is greater than other methods. \vspace{-1.7em}}
	\label{fig:results-kitti-dns1-L}	
\end{figure}

\begin{figure}
	\centering
	\begin{subfigure}{0.45\linewidth}
		\includegraphics[width=\textwidth]{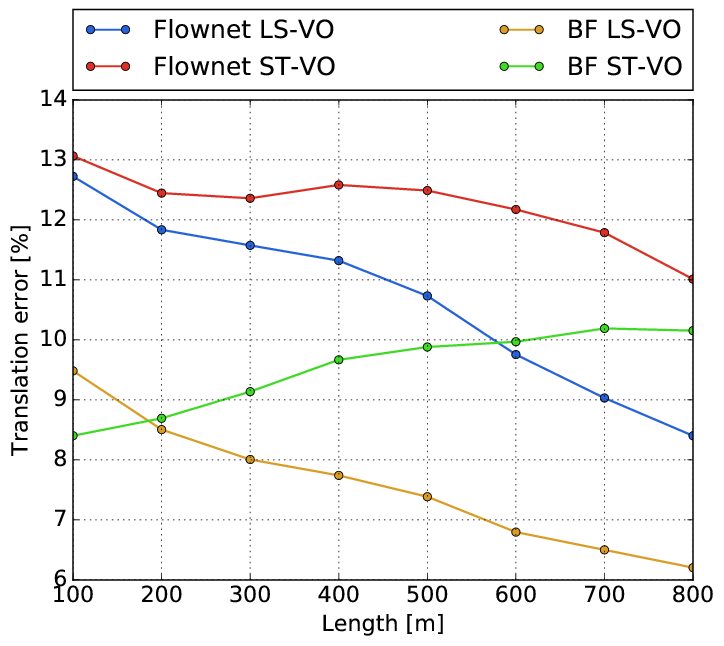}
		\caption{length vs tran. error}
		\label{fig:results2-a}
		\vspace{1mm}
	\end{subfigure}
	\begin{subfigure}{0.45\linewidth}
		\includegraphics[width=\textwidth]{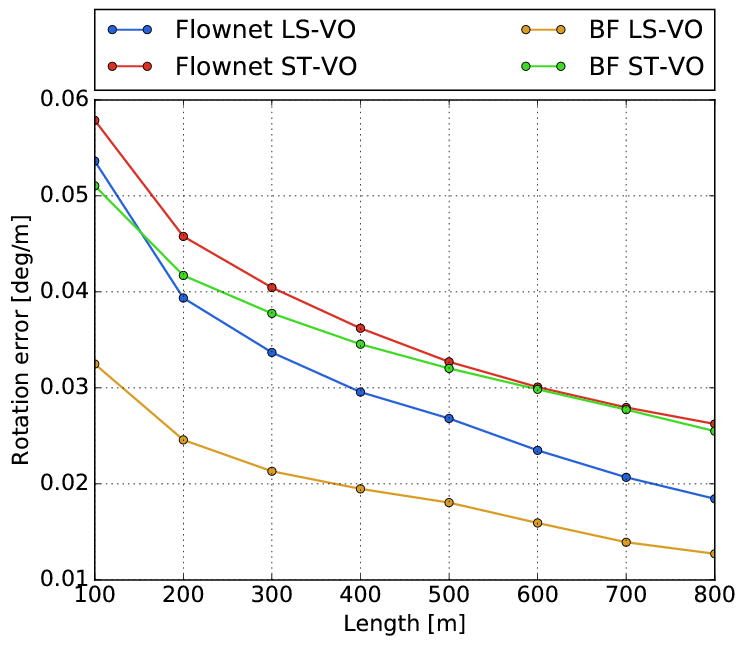}
		\caption{length vs rot. error}
		\label{fig:results2-b}
		\vspace{1mm}
	\end{subfigure}
	\begin{subfigure}{0.45\linewidth}
		\includegraphics[width=\textwidth]{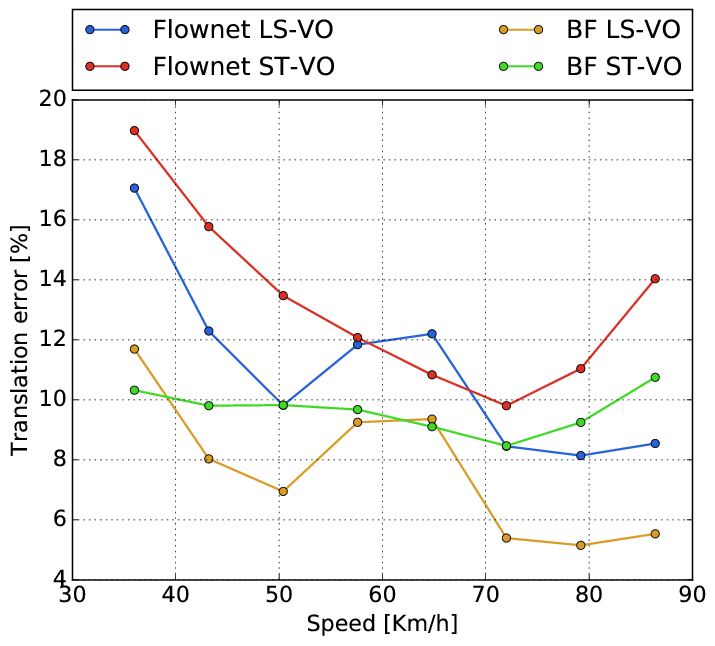}
		\caption{speed vs tran. error}
		\label{fig:results2-c}
	\end{subfigure}
	\begin{subfigure}{0.45\linewidth}
		\includegraphics[width=\textwidth]{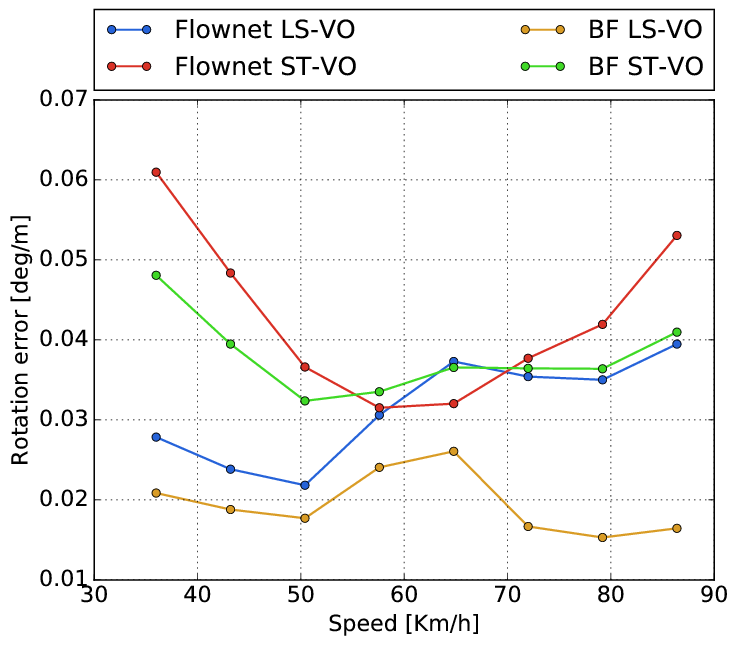}
		\caption{speed vs rot. error}
		\label{fig:results2-d}
	\end{subfigure}
	\caption{\small Comparison between the four network architectures on KITTI \textit{d2} dataset. Again, the LS-VO architecture outperforms the other, except for speed around 60Km/h.\vspace{-1.7em}}
	\label{fig:results2}
\end{figure}

\begin{figure*}[t!]
	\centering
	\begin{subfigure}{0.32\textwidth}
		\includegraphics[width=\linewidth]{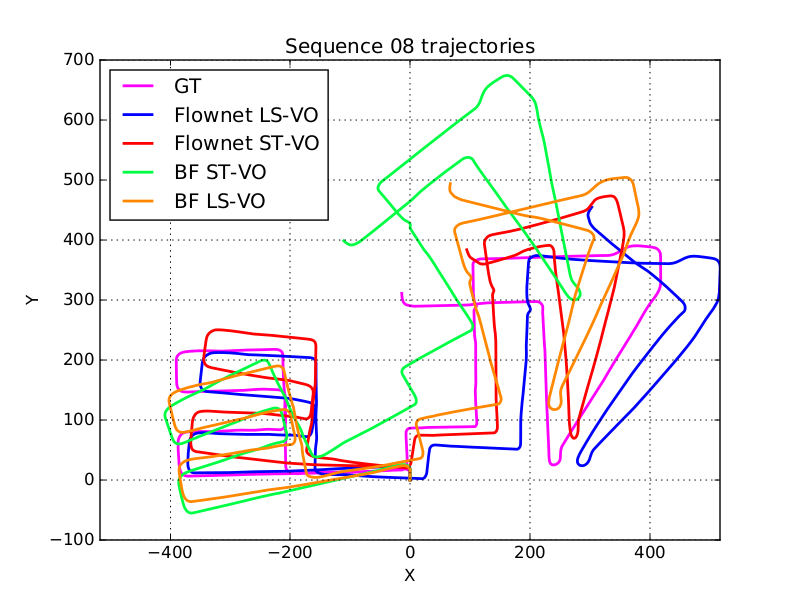}
	\end{subfigure}
	\begin{subfigure}{0.32\textwidth}
		\includegraphics[width=\linewidth]{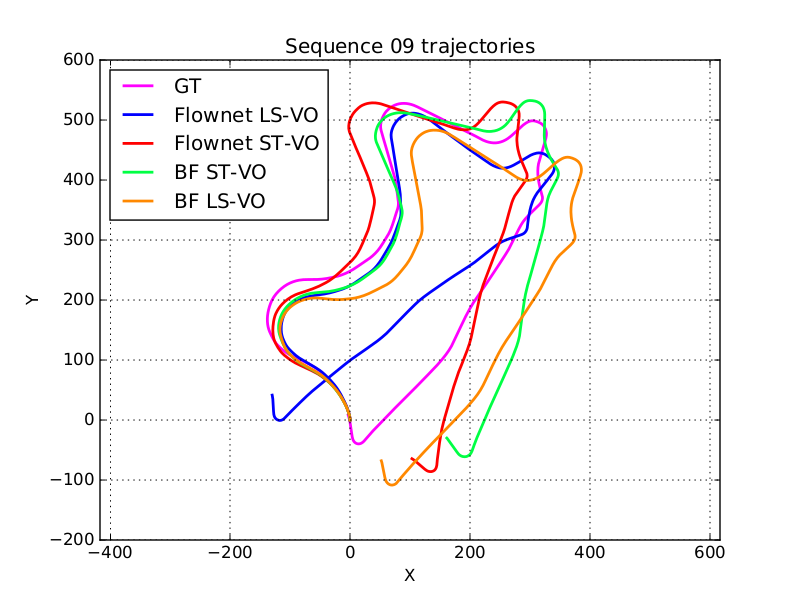}
	\end{subfigure} 
	\begin{subfigure}{0.32\textwidth}
		\includegraphics[width=\linewidth]{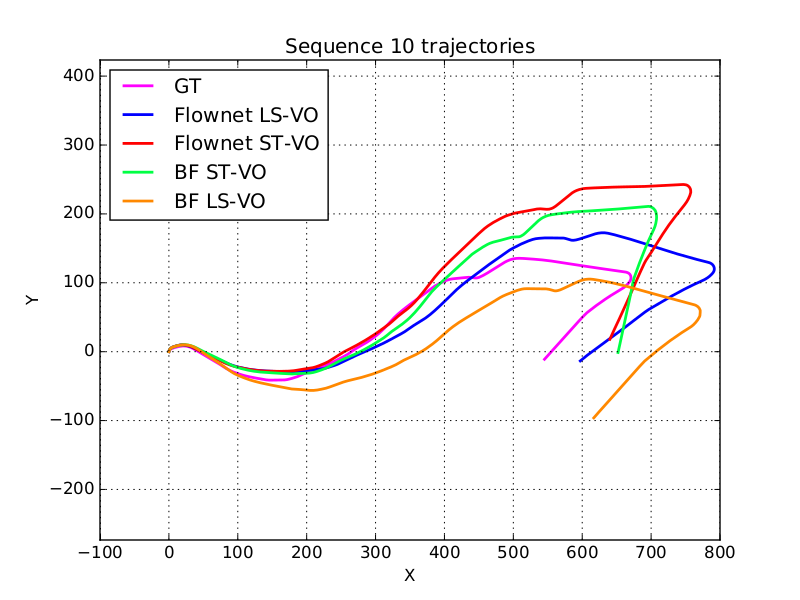}
	\end{subfigure}
	\vspace{-1.0em}
	\caption{\small  KITTI $d2$ trajectories: Trajectories computed on the sub-sampled sequences for all architectures ($d2$ - 5Hz). \vspace{-1.6em}}
	\label{fig:results3}
\end{figure*}


\begin{table*}
\centering

\resizebox{1.0\textwidth}{!}{

\begin{tabular}{|c||c|c|c|c||c|c|c|c||c|c|c|c||}
\hline
 \multirow{2}{*}{} & \multicolumn{2}{c|}{VISO2-M \cite{geiger2011stereoscan}} & \multicolumn{2}{c||}{ORBSLAM2-M \cite{mur2015orb}} & \multicolumn{2}{c|}{ST-VO (Flow)} & \multicolumn{2}{c||}{ST-VO (BF)} & \multicolumn{2}{c|}{LS-VO (Flow)} & \multicolumn{2}{c||}{LS-VO (BF)} \\
\cline{2-13} 
 {} & Trasl. & Rot. & Trasl. & Rot. & Trasl. & Rot. & Trasl. & Rot. & Trasl. & Rot. & Trasl. & Rot.\\
\hline
 KITTI $d1$ & $18.13\%$ & $0.0193$ & $62.71\%$ & $\mathbf{0.0058}$ & $12.73\%$ & $0.0507$ & $8.06\%$ & $0.0205$ & $10.71\%$ & $0.0290$ & $\mathbf{6.98\%}$ & $0.0199$\\
\hline
 KITTI $d2$ & $19.08\%$ & $\mathbf{0.0090}$ & fail & fail & $12.30\%$ & $0.0383$ & $9.43\%$ & $0.0360$ & $10.85\%$ & $0.0320$ & $\mathbf{7.71}\%$ & $0.0205$ \\
\hline
 KITTI $d2$ + blur & $52.54\%$ & $0.0688$ & fail & fail & $18.35\%$ & $0.0502$ & $16.39\%$ & $0.0627$ & $14.47\%$ & $0.0375$ & $\mathbf{8.13\%}$ & $\mathbf{0.02710}$ \\
\hline
\end{tabular}

}

\caption{\small Performances summary of all methods on the Kitti experiments. The geometrical methods perform better on the angular rate estimation (in deg/m) on both datasets at standard rate, but usually fail on others (loss of tracking). Learned methods are consistent in their behaviour in all cases: even if the general error increases, they never fail to give an output even in the worst conditions tested, and the trajectories are always meaningful. \vspace{-1.0em}}
\label{tab:results-kitti-dns1}
\end{table*}

\begin{table*}
\centering
\begin{tabular}{|c||c|c|c|c||c|c|c|c||}
\hline
\multirow{2}{*}{} & \multicolumn{2}{c|}{VISO2-M \cite{geiger2011stereoscan}} & \multicolumn{2}{c||}{ORBSLAM2-M \cite{mur2015orb}} & \multicolumn{2}{c|}{ST-VO (Flow)} & \multicolumn{2}{c|}{LS-VO (Flow)} \\
\cline{2-9} 
 {} & Trasl. & Rot. & Trasl. & Rot. & Trasl. & Rot. & Trasl. & Rot. \\
\hline
 Malaga $d1$ & $43.90\%$ & $0.0321$ & $86.60\%$ & $\mathbf{0.0156}$ & $23.20\%$ & $0.1241$ & $\mathbf{15.56\%}$ & $0.0690$\\
\hline
 Malaga $d2$ & $47.37\%$ & $0.0530$ & fail & fail & $23.35\%$ & $0.1088$ & $\mathbf{21.44}\%$ & $\mathbf{0.0472}$\\
\hline
 Malaga $d2$ + blur & fail & fail & fail & fail & $25.14\%$ & $0.1262$ & $\mathbf{24.06\%}$ & $\mathbf{0.0657}$\\
\hline
\end{tabular}

\caption{\small Performances summary of all methods on the Malaga experiments. The same considerations of Table \ref{tab:results-kitti-dns1} apply. In this set of experiments we analysed only the end-to-end architecture, for the sake of simplicity. \vspace{-1.0em}}
\label{tab:results-malaga-dns1}
\end{table*}

\begin{figure}
	\centering
	\begin{subfigure}{0.45\linewidth}
		\includegraphics[width=\textwidth]{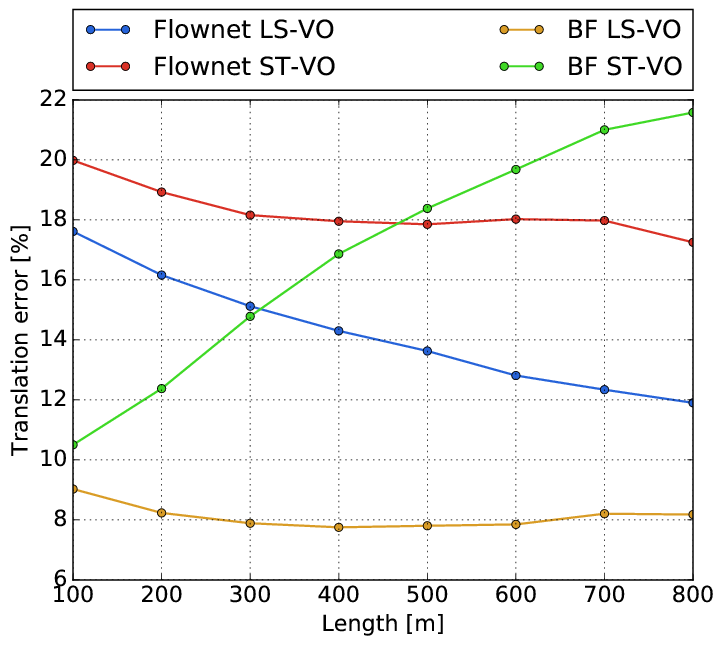}
		\caption{length vs tran. error}
		\label{fig:results4-a}
	\end{subfigure}
	\begin{subfigure}{0.45\linewidth}
		\includegraphics[width=\textwidth]{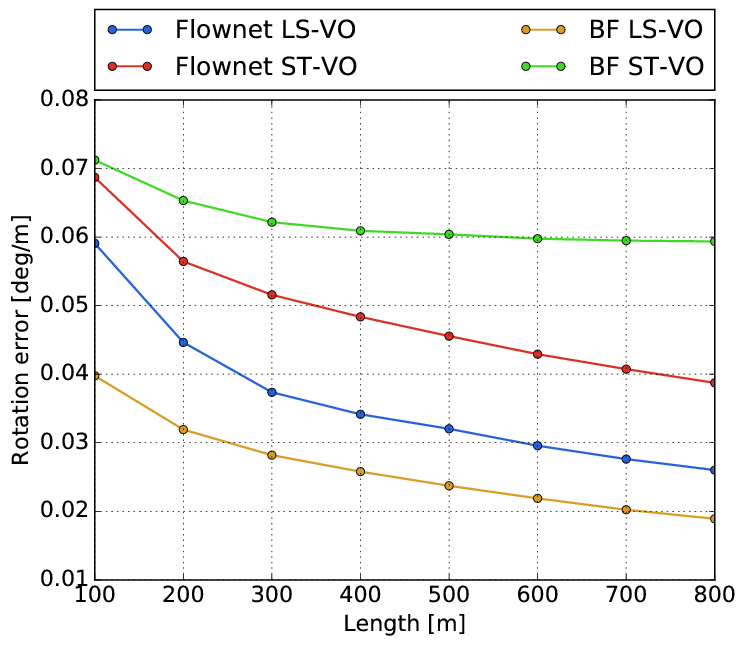}
		\caption{length vs rot. error}
		\label{fig:results4-b}
	\end{subfigure}
	\begin{subfigure}{0.45\linewidth}
		\includegraphics[width=\textwidth]{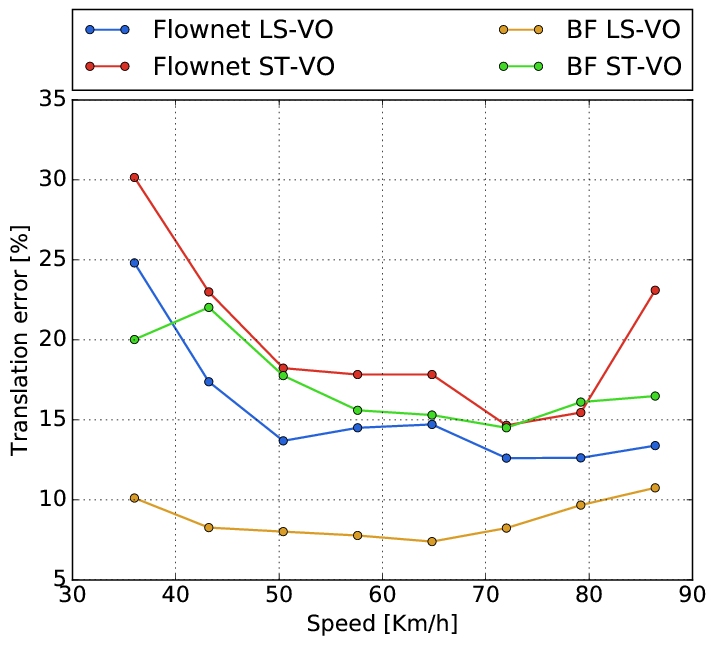}
		\caption{speed vs tran. error}
		\label{fig:results4-c}
	\end{subfigure}
	\begin{subfigure}{0.45\linewidth}
		\includegraphics[width=\textwidth]{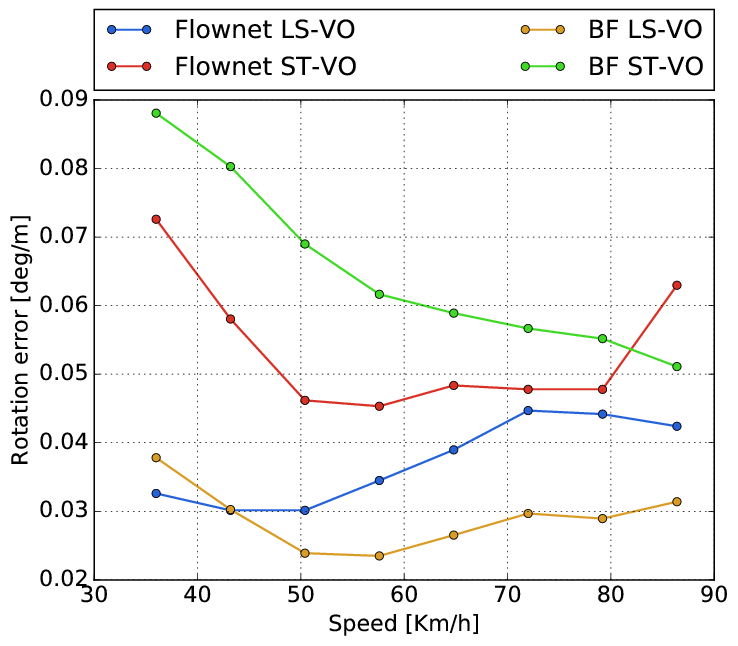}
		\caption{speed vs rot. error}
		\label{fig:results4-d}
	\end{subfigure}
	\caption{\small Performances of the four architectures on blurred KITTI \textit{d2} sequences. The difference in performances between the ST and LS-VO networks is huge. VISO2-M has been omitted, for axis scale reasons. \vspace{-1.6em}}
	\label{fig:results4}
\end{figure}

\begin{figure}
	\centering
	\begin{subfigure}{0.45\linewidth}
		\includegraphics[width=\textwidth]{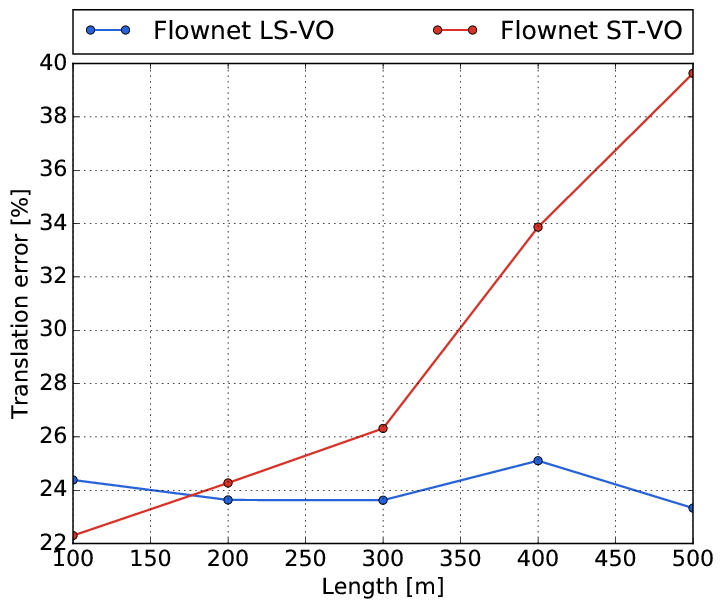}
		\caption{length vs tran. error}
		\label{fig:results4m-a}
	\end{subfigure}
	\begin{subfigure}{0.45\linewidth}
		\includegraphics[width=\textwidth]{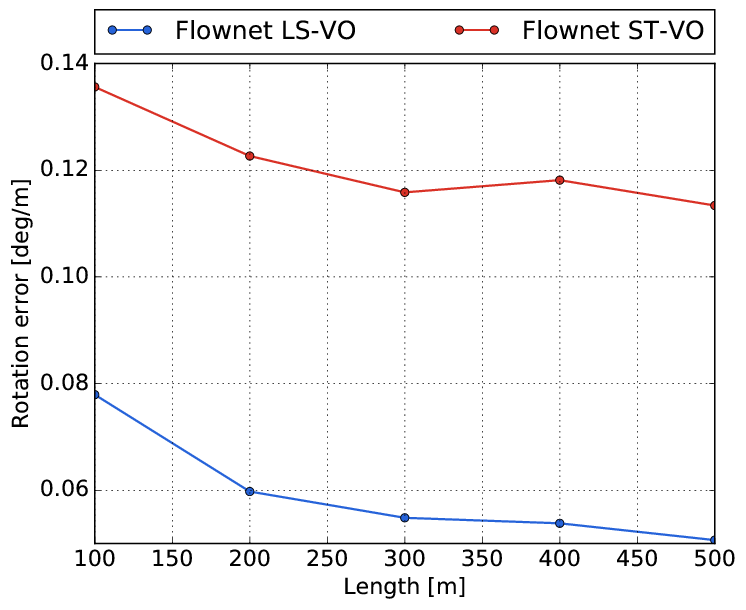}
		\caption{length vs rot. error}
		\label{fig:results4m-b}
	\end{subfigure}
	\begin{subfigure}{0.45\linewidth}
		\includegraphics[width=\textwidth]{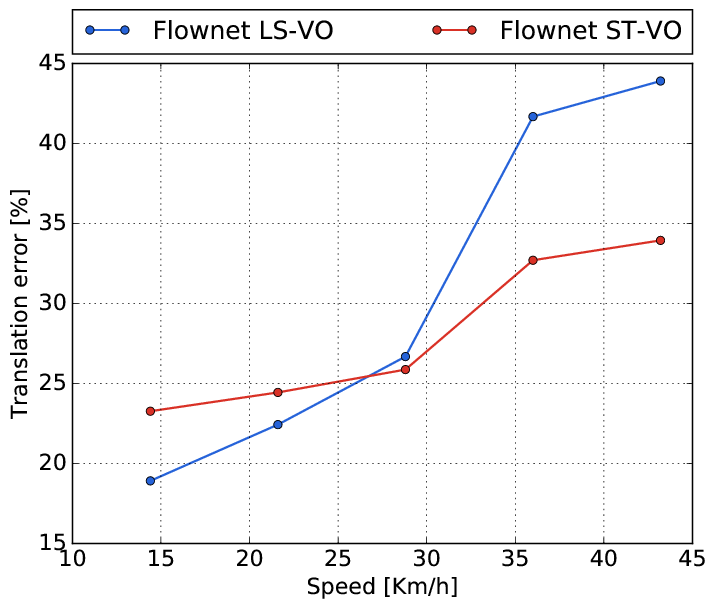}
		\caption{speed vs tran. error}
		\label{fig:results4m-c}
	\end{subfigure}
	\begin{subfigure}{0.45\linewidth}
		\includegraphics[width=\textwidth]{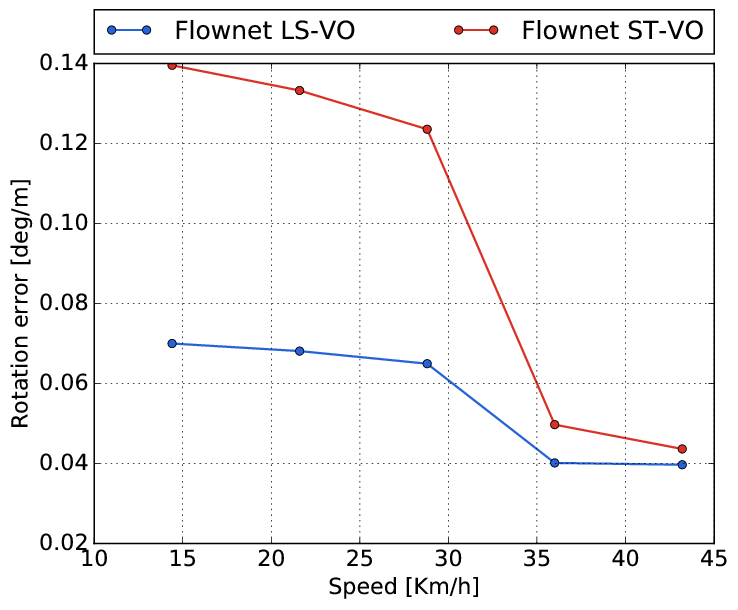}
		\caption{speed vs rot. error}
		\label{fig:results4m-d}
	\end{subfigure}
	\caption{\small  Performances of the end-to-end architecture on blurred Malaga \textit{d2} sequences. The lack of samples at high speed make the LS-VO network slightly overfit those samples, as shown in \ref{fig:results4m-c}, but in all other respects the behaviour is similar to Figure \ref{fig:results4}. }
	\vspace{-1.7em}
	\label{fig:results4m}
\end{figure}

%

\subsection{Discussion}\label{sec:results-experiments}
The experiments described in Section \ref{sec:results-experiments} on both datasets have been evaluated with KITTI devkit \cite{geiger2013vision}, and the output plots have been reported in Figures \ref{fig:results-kitti-dns1-L}, \ref{fig:results2}, \ref{fig:results3}, \ref{fig:results4} and \ref{fig:results4m}. In all Figures except \ref{fig:results3}, the upper sub-plots, (a) and (b), represent the translational and rotational errors averaged on sub-sequences of length $100$m up to $800$m. The lower plots represent the same errors, but averaged on vehicle speed (Km/h). The horizontal axis limits for the lower plots, in Figures relative to $d2$ downsampled experiments are different, since the sub-sampling is seen by the evaluation software as an increase in vehicle speed. In Table \ref{tab:results-kitti-dns1} and \ref{tab:results-malaga-dns1} the total average translational and rotational errors for all the experiments are reported.

Figure \ref{fig:results-kitti-dns1-L} summarises the performances of all methods on KITTI without frame downsampling. From Figures \ref{fig:results-kitti-dns1-L-a} and \ref{fig:results-kitti-dns1-L-b} we observe that the BF-fed architectures outperform the Flownet-fed networks by a good margin. This is expected, since BF OF fields have been tuned on the dataset to be usable, while Flownet has not been fine-tuned on KITTI sequences. In addition, the LS-VO networks perform almost always better than, or on a par with, the corresponding ST networks. When we consider Figures \ref{fig:results-kitti-dns1-L-c} and \ref{fig:results-kitti-dns1-L-d}, we observe that the increase in performance from ST to LS-VO appears to be slight, except in the rotational errors for the Flownet architecture. However, the difference between the length errors and the speed errors is coherent if we consider that the errors are averaged. Therefore, the speed values that are less represented in the dataset are probably the ones that are more difficult to estimate, but at the same time their effect on the general trajectory estimation is consequently less important.

The geometrical methods do not work on frame pairs only, but perform local bundle adjustment and eventually scale estimation. Even if the comparison is not completely fair with respect to learned methods, it is informative nonetheless. In particular we observe (see Figure \ref{fig:results-kitti-dns1-L}) that the geometrical methods are able to achieve top performances on angular estimation, because they work on full-resolution images and because there is no scale error on angular rate. On the contrary, on average, they perform sensibly worse than learned methods for translational errors. This is also expected, since geometrical methods lack in scale estimation, while learned methods are able to infer scale from appearance. Similar results are obtained for the Malaga dataset. The complete set of experiments is available online \cite{online_material}.

When we consider the second type of experiment, we expect that the general performances of all the architectures and methods should decrease, since the task is more challenging. At the same time, we are interested in probing the robustness and generalization properties of the LS-VO architectures over the ST ones. 
Figure \ref{fig:results2} shows the KITTI results. From \ref{fig:results2-a} and \ref{fig:results2-b} we notice that, while all the average errors for each length increase with respect to the previous experiments, they increase much more for the two ST variants.
If we consider the errors depicted in Figures \ref{fig:results2-c} and \ref{fig:results2-d}, we observe that the LS-VO networks perform better than the ST ones, except on speed around 60Km/h, where they are on par. This is understandable, since the networks have been trained on $d1$ and $d3$, that correspond to very low and very high speeds, so the OF in between them are the less represented in the training set. However, the most important consideration here is that the LS-VO architectures show more robustness to domain shifts. The plots of the performances on Malaga can be found online \cite{online_material}, and the same considerations of the previous one apply.

The last experiment is on the downsampled and blurred image. On these datasets both VISO2-M and ORBSLAM2-M fail to give any trajectory, due to the lack of keypoints, while Learned methods always give reasonable results.
The results are shown in Figure \ref{fig:results4} and \ref{fig:results4m} for the KITTI and the Malaga dataset, respectively.
In both KITTI and Malaga experiments we observe a huge improvement in performances of LS-VO over ST-VO. Due to the difference in sample variety in Malaga with respect to KITTI, we observe overfitting of the more complex network (LS-VO) over the less represented linear speeds (above $30$Kmh).  

This experiments demonstrate that the LS-VO architecture is particularly apt to help end-to-end networks in extracting a robust OF representation. This is an important result, since this architecture can be easily included in other end-to-end approaches, increasing the estimation performances by a good margin, but without significantly increasing the number of parameters for the estimation task, making it more robust to overfitting, as mentioned in Section \ref{sec:approach-architecture}.   



\section{Conclusions}
\label{sec:conclusions}

	This work presented LS-VO, a novel network architecture for estimating monocular camera Ego-Motion. The architecture is composed by two branches that jointly learn a latent space representation of the input OF field, and the camera motion estimate. The joint training allows for the learning of OF features that take into account the underlying structure of a lower dimensional OF manifold. 
	The proposed architecture has been tested on the KITTI and Malaga datasets, with challenging alterations, in order to test the robustness to domain variability in both appearance and OF dynamic range. Compared to the data-driven architectures, LS-VO network outperformed the single branch network on most benchmarks, and in the others performed at the same level. Compared to geometrical methods, the learned methods show outstanding robustness to non-ideal conditions and reasonable performances, given that they work only on a frame to frame estimation and on smaller input images.
	The new architecture is lean and easy to train and shows good generalization performances. The results provided here are promising and encourage further exploration of OF field latent space learning for the purpose of estimating camera Ego-Motion. 
	All the code, datasets and trained models are made available online \cite{online_material}.


\bibliographystyle{IEEEtran}
\bibliography{bibliography}

\end{document}